\documentclass[runningheads,a4paper]{llncs}

\usepackage{amssymb}
\usepackage{graphicx}
\usepackage{float}

\usepackage{url}

\begin{document}

\mainmatter  

\title{Temporal Convolution Networks for Real-Time Abdominal Fetal Aorta Analysis with Ultrasound}

\author{Nicol\'o Savioli$^1$, Silvia Visentin$^{2}$, Erich Cosmi$^{2}$, Enrico Grisan$^{1,3}$, Pablo Lamata$^1$, Giovanni Montana$^{1,4}$\\
$^1$ Department of Biomedical Engineering, King’s College London, SE1 7EH, UK, \{nicolo.l.savioli,enrico.grisan,pablo.lamata,giovanni.montana\}@kcl.ac.uk \\
$^2$ Department of Woman and Child Health,University Hospital of Padova, Padova, Italy, \{silvia.visentin.1,erich.cosmi\}@unipd.it \\
$^3$ Department of Information Engineering, University of Padova, Padova, Italy, enrigri@dei.unipd.it \\
$^4$ WMG, University of Warwick, Coventry, CV4 71AL, g.montana@warwick.ac.uk}


%
%

%

\institute{}

%
%

\toctitle{Lecture Notes in Computer Science}
\tocauthor{}
\maketitle

\begin{abstract}
The automatic analysis of ultrasound sequences can substantially improve the efficiency of clinical diagnosis.
In this work we present our attempt to automate the challenging task of measuring the vascular diameter of the fetal abdominal
aorta from ultrasound images. We propose a neural network 
architecture consisting of three blocks: a convolutional layer for the extraction of imaging features, a Convolution Gated Recurrent Unit (C-GRU) for enforcing the temporal coherence across video frames and exploiting the temporal redundancy of a signal, and a regularized loss function, called \textit{CyclicLoss}, to impose our prior knowledge about the periodicity of the observed signal. 
We present experimental evidence suggesting that the proposed architecture can reach an accuracy substantially superior to previously proposed methods, providing an average reduction of the mean squared error from $0.31 mm^2$ (state-of-art) to $0.09 mm^2$, and a relative error reduction from $8.1\%$ to $5.3\%$. The mean execution speed of the proposed approach of 289 frames per second makes it suitable for real time clinical use.
\end{abstract}


\section{Introduction}

Fetal ultrasound (US) imaging plays a fundamental role in the monitoring of fetal growth during pregnancy and in the measurement of the fetus well-being. Growth monitoring is becoming increasingly important since there is an epidemiological evidence that abnormal birth weight is associated with an increased predisposition to diseases related to cardiovascular risk (such as diabetes, obesity, hypertension) in young and adults \cite{Visentin14}. 

Among the possible biomarkers of adverse cardiovascular remodelling in fetuses and newborns, the most promising ones are the Intima-Media Thickness (IMT) and the stiffness of the abdominal aorta by means of ultrasound examination. Obtaining reliable measurements is critically based on the accurate estimation of the diameter of the aorta over time.
However, the poor signal to noise ratio of US data and the fetal movement makes the acquisition of a clear and stable US video challenging. Moreover, the measurements rely either on visual assessment at bed-side during patient examination, or on tedious, error-prone and operator-dependent review of the data and manual tracing at later time. Very few attempts towards automated assessment have been presented \cite{VeroneseE,Tarroni15}, all of which have computational requirements that prevent them to be used in real-time. As such, they have reduced appeal for the clinical use.
In this paper we describe a method for automated measurement of the abdominal aortic diameter directly from fetal US videos. We propose neural network architecture that is able to process US videos in real-time and leverage both the temporal redundancy of US videos and the quasi-periodicity of the aorta diameter.

The main contributions of the proposed method are as follows. First we show that a shallow CNN is able to learn imaging features and outperforms classical methods as level-set for fetal abdominal aorta diameter prediction. Second we add to the CNN a Convolution Gated Recurrent Unit (C-GRU) \cite{Mennatullah} for exploiting the temporal redundancy of the features extracted by CNN from the US video sequence. Finally, we add a new penalty term to the loss function used to train the CNN to exploit periodic variations. 

\section{Related work}

The interest for measuring the diameter and intima-media thickness (IMT) of major vessels has stemmed from its importance as biomarker of hypertension damage and atherosclerosis in adults. Typically, the IMT is assessed on the carotid artery by identifying its lumen and the different layers of its wall on high resolution US images. The improvements provided by the design of semi-automatic and automatic methods based mainly on the image intensity profile, distribution and gradients analysis, and more recently on active contours. For a comprehensive review of these classical methods we refer the reader to \cite{Molinari} and \cite{Loizou}. 
In the prenatal setting, the lower image quality, due to the need of imaging deeper in the mother's womb and by the movement of the fetus, makes the measurement of a IMT biomarker, although measured on the abdominal aorta, challenging.

Methods that proved successful for adult carotid image analysis do not perform well on such data, for which only a handful of methods (semi-automatic or automatic) have been proposed, making use of classical tracing methods and mixture of Gaussian modelling of blood-lumen and media-adventitia interfaces \cite{VeroneseE}, or on level sets segmentation with additional regularizing terms linked to the specific task \cite{Tarroni15}. However, their sensitivity to the image quality and lengthy computation prevented an easy use in the clinical routine.

Deep learning approaches have outperformed classical methods in many medical tasks \cite{Geert}. The first attempt in using a CNN, for the measurement of carotid IMT has been made only recently \cite{JYShin}. In this work, two separate CNNs are used to localize a region of interest and then segment it to obtain the lumen-intima and media-adventitia regions.
Further classical post-processing steps are then used to extract the boundaries from the CNN based segmentation. The method assumes the presence of strong and stable gradients across the vessel walls, and extract from the US sequence only the frames related to the same cardiac phase, obtained by a concomitant ECG signal. 

However, the exploitation of temporal redundancy on US sequences was shown to be a solution for improving overall detection results of the fetal heart \cite{WeilinHuang}, where the use of a CNN coupled with a recurrent neural network (RNN) is strategic. Other works, propose similar approach in order to detect the presence of standard planes from prenatal US data using CNN with Long-Short Term Memory (LSTM) \cite{Chen2}.

\section{Datasets}

This study makes use of a dataset consisting of 25 ultrasound video sequences acquired during routine third-trimester pregnancy check-up at the Department of Woman and Child Health of the University Hospital of Padova (Italy). The local ethical committee approved the study and all patients gave written informed consent.

Fetal US data were acquired using a US machine (Voluson E8, GE) equipped with a 5 MHz linear array transducer, according to the guidelines in \cite{Cosmi,Skilton},
using a $70^o$ FOV, image dimension  720x960  pixels, a  variable  resolution  between 0.03 and 0.1 $mm$ and a mean frame rate of 47 fps. Gain settings were tuned to enhance the visual quality and contrast during the examination. The length of the video is between 2s and 15s, ensuring that at least one full cardiac cycle is imaged.

After the examination, the video of each patient was reviewed and a relevant video segment was selected for semi-automatic annotation considering its visual quality and length: all frames of the segment were processed with the algorithm described in \cite{VeroneseE} and then the diameters of all frames in the segments were manually reviewed and corrected. The length of the selected segments varied between 21 frames 0.5s and 126 frames 2.5s. 
The 25  annotated segments in the dataset were then randomly divided into training ($60\%$ of the segments), validation ($20\%$) and testing ($20\%$) sets.
In order to keep the computational and memory requirements low, each frame was cropped to have a square aspect ratio and then resized to $128 \times 128$ pixels. We also make this dataset public to allow the results reproducibility.

\section{Network architecture}

Our output is the predicted value $\hat{y}[t]$ of the diameter of the abdominal aorta at each time point. 
Our proposed deep learning solution consists of three main components (see Figure [\ref{fig:fig1}]): a Convolutional Neural Network (CNN) that captures the salient characteristics from ultrasound input images; a Convolution Gated Recurrent Unit (C-GRU) \cite{Mennatullah} exploits the temporal coherence through the sequence; and a regularized loss function, called \textit{CyclicLoss}, that exploits the redundancy between adjacent cardiac cycles. 

Our input consists of a set of sequences whereby each sequence $S=[s[1], ..., s[K]]$ has dimension  $N \times M$ pixels at time $t$, with $t\in\{1,\ldots,K\}$. At each time point t, the CNN extracts the feature maps $x[t]$ of dimensions $D\times N_x \times M_x$, where $D$ is the number of maps, and $N_x$ and $M_x$ are their in-plane pixel dimensions, that depend on the extent of dimensionality reduction obtained by the CNN through its pooling operators.

The feature maps are then processed by a C-GRU layer \cite{Mennatullah}. The C-GRU combines the current feature maps $x[t]$ with an encoded representation $h[t-1]$ of the feature maps $\{x[1],\ldots,x[t-1]\}$ extracted at previous time points of the sequence to obtain an updated encoded representation $h[t]$, the \textit{current state}, at time $t$: this allows to exploit the temporal coherence in the data. The $h[t]$ of the C-GRU layer is obtained by two specific gates designed to control the information inside the unit: a reset gate, $r[t]$, and an update gate, $z[t]$, defined as follow:
\begin{equation}
r[t]  = \sigma (W_{hr} * h[t-1] + W_{xr}* x[t] + b_{r}) \\
z[t]  = \sigma (W_{hz} * h[t-1] + W_{xz}* x[t] + b_{z})
\end{equation}
Where, $\sigma ()$ is the sigmoid function, $W_{\cdot}$ are recurrent weights matrices whose first subscript letter refers to the input of the convolution operator (either the feature maps $x[t]$ or the state $h[t-1]$), and whose second subscript letter refers to the gate (reset $r$ or update $z$). 
All this matrices, have a dimension of $D \times 3 \times 3$ and $b_{\cdot}$ is a bias vector. 
In this notation, $*$ defines the convolution operation. The current state is then obtained as:
\begin{equation}
h[t] = (1-z[t]) \odot  h[t-1] + z[t] \odot \tanh(W_{h}*(r[t] \odot h_{t-1}) + W_{x} *x[t] + b).
\end{equation}
Where $\odot$ denotes the dot product and $W_{h}$ and $W_{x}$ are recurrent weight matrices for $h[t-1]$ and $x[t]$, used to balance the new information represented by the feature maps $x[t]$ derived by the current input data $s[t]$ with the information obtained observing previous data $s[1],\ldots,s[t-1]$.
On the one hand, $h[t]$ is then passed on for updating the state $h[t+1]$ at the next time point, and on the other is flatten and fed into the last part of the network, built by Fully Connected (FC) layers progressively reducing the input vector to a scalar output that represent the current diameter estimate $\hat{y}[t]$.
\begin{figure*}[ht]
\centering
 \includegraphics[width=5in]{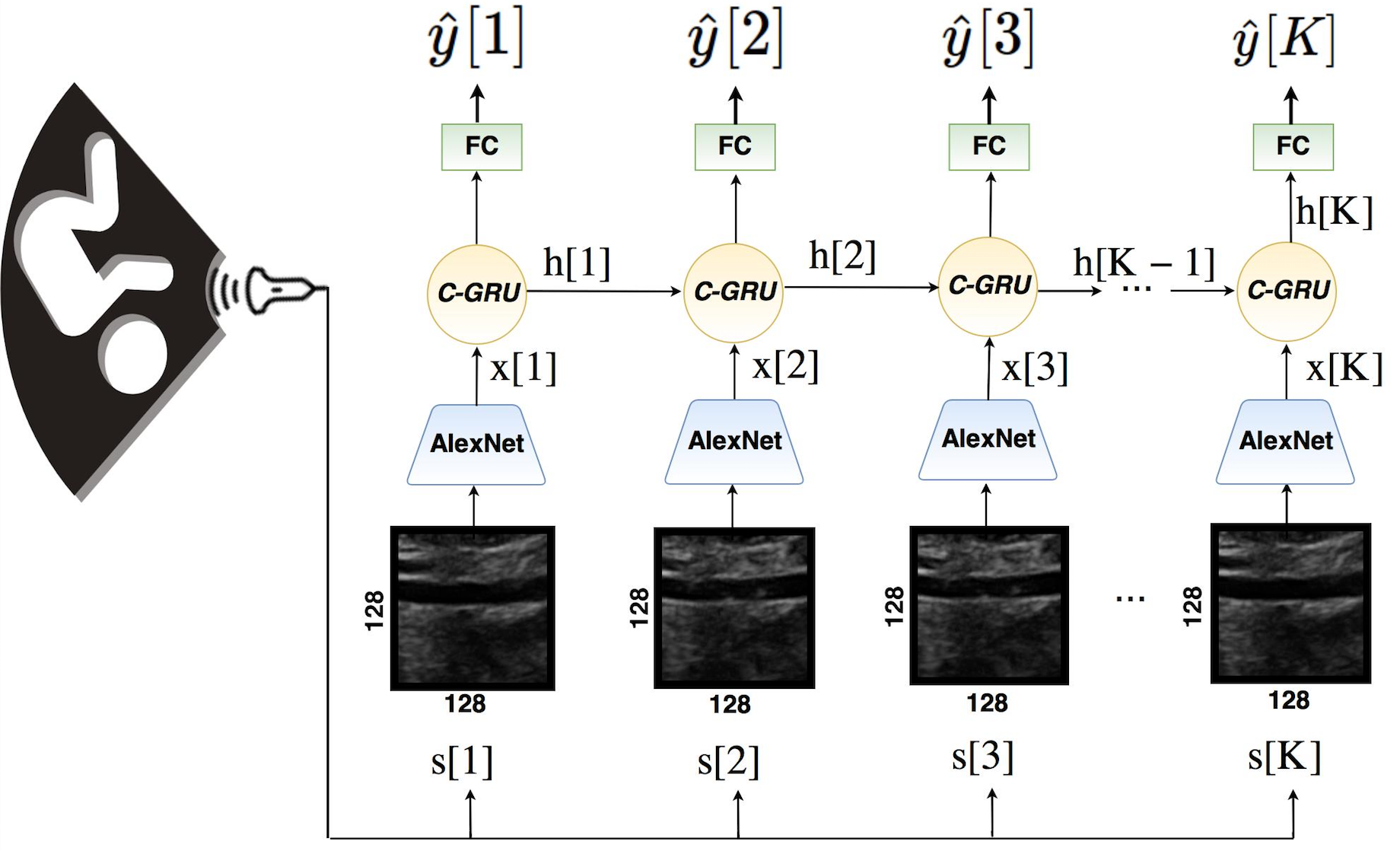}
   \caption{The deep-learning architecture proposed for abdominal diameter aorta prediction. The blue blocks represent the features extraction through a CNN (AlexNet) which takes in input a US sequence $S$, and provides for each frame $s[t]$ a features map $x[t]$ that is passed to Convolution Gated Recurrent Units (C-GRU) (yellow circle) that encodes and combines the information from different time points to exploit the temporal coherence. The fully connected block (FC, in green), takes as input the current encoded state $h[t]$ as features to estimate the aorta diameter $\hat{y}[t]$.}
\label{fig:fig1}
\end{figure*}

\subsection{CyclicLoss}

Under the assumption that the pulsatility of the aorta follows a periodic pattern with the cardiac cycle, the diameter of the vessel at corresponding instants of the cardiac cycle should ideally be equal. Assuming a known cardiac period $T_{period}$, we propose to add a regularization term to the loss function used to train the network as to penalize large differences of the diameter values that are estimated at time points that are one cardiac period apart.

We call this regularization term \textit{CyclicLoss} ($CL$), computed as $L_2$ norm between pairs of predictions at the same point of the heart cycle and from adjacent cycles:

\begin{equation}
CL=\sqrt{\sum_{n=1}^{N_{cycles}} \sum_{t=0}^{T_{period}} \parallel \hat{y}[t+(n-1)T_{period}] - \hat{y}[t+nT_{period}]  \parallel_{2}}
\end{equation}

The $T_{period}$ is the period of the cardiac cycle, while $N_{cycles}$ is the number of integer cycles present in the sequence and $\hat{y}[t]$ is the estimated diameter at time $t$.
Notably, the $T_{period}$ is determined through a peak detection algorithm on $y[t]$, and the average of all peak-to-peak detection distances define its value. While the $N_{cycles}$ is the number of cycles present, calculated as the total length of the $y[t]$ signal divided by $T_{period}$.

The loss to be minimized is therefore a combination of the classical mean squared error (MSE) with the $CL$, and the balance between the two is controlled by a constant $\lambda$: 
\begin{equation}
 Loss = MSE + \lambda\cdot CL = \frac{1}{K} \sum_{t=1}^{K} (y[t]-\hat{y}[t])^{2} + \lambda\cdot CL
\end{equation}
where $y[t]$ is the target diameter at time point $t$.
It is worth noting that the knowledge of the period of the cardiac cycle is needed only during training phase. Whereas, during the test phase, on unknown image sequence, the trained network provide its estimate blind of the periodicity of the specific sequence under analysis.
 
\begin{figure}[h]
\centering
 \includegraphics[width=12cm,height=10cm,scale=0.6]{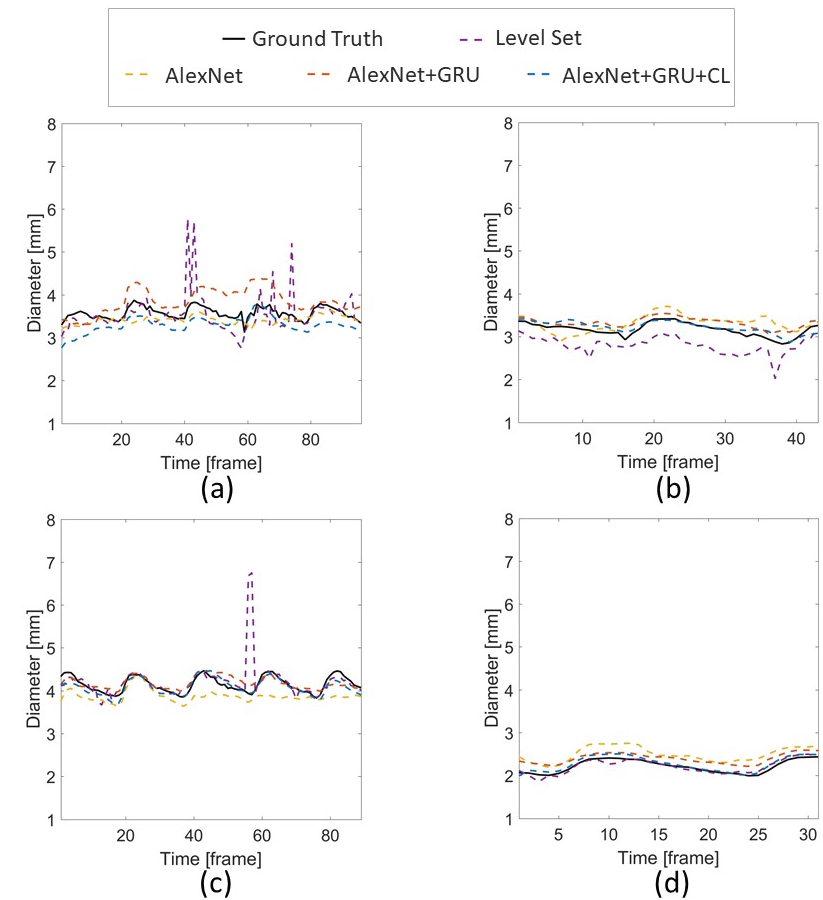}
   \caption{Each panel (a-c) shows the estimation of the aortic diameter at each frame of fetal ultrasound videos in the test set, using the level set method (dashed purple line), the naive architecture using AlexNet (dashed orange line), the AlexNet+C-GRU (dashed red line), and AlexNet+C-GRU trained with the \textit{CyclicLoss} (dashed blue line). The ground truth (solid black line) is reported for comparison. Panels (a,c) show the results on long sequences where more than 3 cardiac cycles are imaged, whereas panels (b,d) show the results on short sequences where only 1 or two cycles are available.}
  \label{fig:fig1}
\end{figure}

\subsection{Implementation details}

For our experiments, we chose AlexNet \cite{AlexKrizhevsky} as a feature extractor for its simplicity. It has five hidden layers with $11 \times 11$ kernels size in the first layer, $5 \times 5$ in the second and $3 \times 3$ in the last three layers; it is well suited to the low image contrast and diffuse edges characteristic of US sequences.
Each network input for the training is a sequence of $K=125$ ultrasound frames with $N=M=128$ pixels, AlexNet provides feature maps of dimension $D\times N \times M=256\times 13 \times 13$, and the final output $\hat{y}[t]$ is the estimate abdominal aorta diameter value at each frame.

The loss function is optimised with the Adam algorithm \cite{Kingma2014AdamAM} that is a first-order gradient-based technique. The learning rate used is $1e^{-4}$ with $2125$ iterations (calculated as number of patients $\times$ number of ultrasound sequences) for $100$ epochs. In order to improve generalization, data augmentation of the input with a vertical and horizontal random flip is used at each iteration. The $\lambda$ constant used during training with 
\textit{CyclicLoss} takes the value
of $1e^{-6}$.

\section{Experiments}

The proposed architecture is compared with the currently adopted approach in section 4. This method provides fully-automated measurements in lumen identification on prenatal US images of the abdominal aorta \cite{Tarroni15} based on edge-based level set. In order to understand the behaviour of different features extraction methods, we have also explored the performance of new deeper network architectures whereby AlexNet was replaced it by InceptionV4 \cite{ChristianSzegedy} and DenseNets 121 \cite{GaoHuang}.
\begin{table}[]
\centering
\label{my-label}
\begin{tabular}{|l|l|l|l|l|}
\hline
\textbf{Methods}                   & \textbf{MSE [$mm^2$]}     & \textbf{RE [\%]}   & \textbf{p-value} \\ \hline
AlexNet                            &  0.29(0.09)              & 8.67(10)            &  1.01e-12         \\ \hline
AlexNet+C-GRU                      &  0.093(0.191)            & 6.11(5.22)          &  1.21e-05          \\ \hline
\textbf{AlexNet+C-GRU+CL}          & \textbf{0.085(0.17)}     & \textbf{5.23(4.91)} &  ``-'' 			    \\ \hline
DenseNet121                        &  0.31(0.56)              & 9.55(8.52)          &  6.00e-13 			 \\ \hline
DenseNet121+C-GRU                  &  0.13(0.21)              & 7.72(5.46)          &  7.78e-12  		      \\ \hline
InceptionV4                        &  6.81(14)                & 50.4(39.5)          &  6.81e-12  			   \\ \hline
InceptionV4+C-GRU                  &  0.76(1.08)              & 16.3(9.83)          &  2.89e-48      			\\ \hline
Level-set                          &  0.31(0.80)              & 8.13(9.39)          &  1.9e-04    	         \\ \hline
\end{tabular}
\vspace{2mm}
\caption{The table show the mean (standard deviation) of MSE and RE error for all the comparison models. The combination of C-GRU and the \textit{CyclicLoss} with AlexNet yields the best performance. Adding recurrent units to any CNN architecture improves its performance; however deeper networks as InceptionV4 and DenseNets do not show any particular benefits with respect to the simpler AlexNet.
Notably, we also consider the p-value for multiple models comparison with the propose network AlexNet+C-GRU+CL, in this case the significant level should be 0.05/7 using the Bonferroni correction \cite{Bonferroni}.
}
\label{table:table1}
\end{table}

The performance of each method was evaluated both with respect to the mean squared error (MSE) and to the mean absolute relative error (RE); all values are reported in Tab.\ref{table:table1} in terms of average and standard deviation across the test set.

In order to provide a visual assessment of the performance, representative estimations on four sequences of the test set are shown in Fig.\ref{fig:fig1}. The naive architecture relying on a standard loss and its C-GRU version are incapable to capture the periodicity of the diameter estimation. The problem is mitigated by adding the \textit{CyclicLoss} regularization on MSE. This is quantitatively shown in Tab.\ref{table:table1}, where the use of this loss further decreases the MSE from $0.093 mm^2$ to $0.085 mm^2$, and the relative error of from $6.11 \%$ to $5.23 \%$. 

Strikingly, we observed that deeper networks are not able to outperform AlexNet on this dataset. Their limitation may be due to over-fitting. Nevertheless, the use of C-GRU greatly improve the performance of both networks both in terms of MSE and of RE. 
Further, we also performed a non-parametric test
(Kolmogorov-Smirnov test) to check if the best model was statistically different compared to the others.

The results obtained with the complete model AlexNet+C-GRU+CL are indeed significantly different from all others (p $\textless$ 0.05) also, when the significant level is adjusted for multiple comparison applying the Bonferroni correction \cite{Bonferroni,Dunn}.

\section{Discussion and conclusion}

The deep learning (DL) architecture proposed shows excellent performance compared to traditional image analysis methods, both in accuracy and efficiency. This improvement is achieved through a combination of a shallow CNN and the exploitation of the temporal and cyclic coherence. Our results seem to indicate that a shallow CNNs perform better than deeper CNNs such as DenseNet 121 and InceptionV4; this might be due to the small dimension of the data set, a common issue in the medical settings when requiring manual annotations of the data.  

\subsection{The \textit{CyclicLoss} benefits}

The exploitation of temporal coherence is what pushes the performance of the DL solution beyond current image analysis methods, reducing the MSE from $0.29 mm^2$ (naive architecture) to $0.09 mm^2$ with the addition of the C-GRU. The \textit{CyclicLoss} is an efficient way to guide the training of the DL solution in case of data showing some periodicity, as in cardiovascular imaging. 
Please note that the knowledge of the signal period is only required by the network during training, and as such it does not bring additional requirements on the input data for real clinical application.
We argue that the \textit{CyclicLoss} is making the network learn to expect a periodic input and provide some periodicity in the output sequence. 

\subsection{Limitations and future works}

A drawback of this work is that it assumes the presence of the vessel in the current field of view. 
Further research is thus required to evaluate how well the solution adapts to the scenario of lack of cyclic consistency,  when the vessel of interest can move in and out of the field of view during the acquisition, and to investigate the possibility of a concurrent estimation of the cardiac cycle and vessel diameter. 
Finally, the C-GRU used in our architecture, has two particular advantages compared to previous approaches \cite{Chen2,WeilinHuang}: first, it is not subject to the vanishing gradient problem as the RNN, allowing to train from long sequences of data.
Second, it has less computational cost compared to the LSTM, and that makes it suitable for real time video application. 

\bibliographystyle{plain}

\end{document}